\documentclass{article}

\PassOptionsToPackage{numbers, compress}{natbib}
    \usepackage[final]{neurips_2019}

\usepackage[utf8]{inputenc} % allow utf-8 input
\usepackage[T1]{fontenc}    % use 8-bit T1 fonts
\usepackage{hyperref}       % hyperlinks
\usepackage{url}            % simple URL typesetting
\usepackage{booktabs}       % professional-quality tables
\usepackage{amsfonts}       % blackboard math symbols
\usepackage{nicefrac}       % compact symbols for 1/2, etc.
\usepackage{microtype}      % microtypography
\usepackage{graphicx}
\usepackage{amsmath}
\usepackage{multicol}
\usepackage{xcolor}
\usepackage{mwe}
\usepackage{float} % For specifying table/figure locations, i.e. [ht!]
\usepackage{ragged2e} %justify text

\usepackage{lipsum}
\usepackage{epsfig}
\usepackage{amsmath}
\usepackage{amssymb}
\usepackage{multicol}
\usepackage{graphicx}
\usepackage{booktabs}       % professional-quality tables
\usepackage[first=0,last=9]{lcg}
\usepackage{color, colortbl}
\usepackage{xcolor}
\usepackage{commath}

\title{Deep Bayesian Recurrent Neural Networks for Somatic Variant Calling in Cancer}

\author{
  Geoffroy Dubourg-Felonneau$^1$, Omar Darwish$^{1,2}$, Christopher Parsons$^1$, D\`{a}mi Rebergen$^1$,\\
  \textbf{John W Cassidy$^1$, Nirmesh Patel$^1$, Harry W Clifford$^1$} \\
  \\
  \and
  $^1$Cambridge Cancer Genomics \\
  Cambridge, UK \\
  www.ccg.ai
  \and
  $^2$The University of Cambridge \\
  Cambridge, UK \\
}

\begin{document}

\maketitle

\begin{abstract}

The emerging field of precision oncology relies on the accurate pinpointing of alterations in the molecular profile of a tumor to provide personalized targeted treatments. Current methodologies in the field commonly include the application of next generation sequencing technologies to a tumor sample, followed by the identification of mutations in the DNA known as somatic variants. The differentiation of these variants from sequencing error poses a classic classification problem, which has traditionally been approached with Bayesian statistics, and more recently with supervised machine learning methods such as neural networks. Although these methods provide greater accuracy, classic neural networks lack the ability to indicate the confidence of a variant call. In this paper, we explore the performance of deep Bayesian neural networks on next generation sequencing data, and their ability to give probability estimates for somatic variant calls. In addition to demonstrating similar performance in comparison to standard neural networks, we show that the resultant output probabilities make these better suited to the disparate and highly-variable sequencing data-sets these models are likely to encounter in the real world. We aim to deliver algorithms to oncologists for which model certainty better reflects accuracy, for improved clinical application. By moving away from point estimates to reliable confidence intervals, we expect the resultant clinical and treatment decisions to be more robust and more informed by the underlying reality of the tumor molecular profile.

\end{abstract}

\section{Introduction}

Cancer is a disease of the genome, in which a range of structural genomic changes to the DNA of otherwise healthy cells, results in a cancer phenotype. A well-studied subset of these aberrations is single nucleotide variants and short insertion/deletion mutations. Identification of these through next generation sequencing for a given individual can now be commonly used in clinical practice to provide targeted and personalized therapies; a field known as precision oncology \cite{Bungartz2018MakingTR}. However, the deconvolution of these somatic variants from sequencing error provides a somatic/artifact classification problem which is yet to be fully resolved.

This problem has been traditionally approached with Bayesian classifiers \cite{Cibulskis2013SensitiveDO, Saunders2012StrelkaAS}, which give reasonable accuracy whilst providing information on the robustness of the variant call. More recently, there have been attempts to increase the accuracy of variant calls through neural network-based approaches, such as DeepVariant (germline variants)\cite{Poplin2018AUS} and SomaticNET\cite{ESMOnet,Harries2018} (somatic variants). However, these are yet to provide the advantage of output probabilities that better represents the underlying reality of the tumor.

In this study, we aim to build upon our previous work in applying Bayesian Neural Networks (BNNs) to somatic variant calling \cite{DubourgFelonneau2019SafetyAR}, an approach which enables the network to learn a posterior probability density function (pdf) over the space $\mathcal{W}$ of the weights, and in turn provide a measure of uncertainty which makes the model more robust. We build on this by investigating the distribution of output probabilities compared to standard neural networks, when applied to real-world data and in various out-of-distribution scenarios commonly observed in disparate sequencing data-sets.

\section{Variant Calling}

The task, data \& labels for this study are consistent with our previous work in this field\cite{DubourgFelonneau2019SafetyAR} as follows:

\subsection{The Task}

Next generation DNA sequencing techniques produce overlapping short sequences of DNA called reads. For a fixed position in a sequenced genome, we observe a number of reads originating from multiple cells. We call this number the depth. When, for a given patient, we sequence both tumor and normal tissues for comparison, we expect to see a number of true somatic variants (in the  tumor tissue but not in the normal tissue), although the vast majority of the variations observed are due to sequencing noise. The task we undertake here involves the differentiation of somatic variants (tumor) from germline variants (normal) and sequencing error (noise) .

\subsection{The Data}

We developed a bioinformatics pipeline to identify genomic loci where sequencing data from the tumor significantly differs from the normal. We extract the data in the following form:

\begin{multicols}{2}
        \[
        pair
        =
        \left[
        \begin{array}{ccc|ccc}
            N_{11} & \dots  & N_{1w}  &  T_{11} & \dots  & T_{1w} \\
            \vdots & \ddots & \vdots  &  \vdots & \ddots & \vdots \\
            N_{d1} & \dots  & N_{dw}  &  T_{d1} & \dots  & T_{dw} \\
        \end{array}
        \right]
        \]
        
        $d$ : the depth \\
        $w$ : the width or number of observed loci \\
        $N_{ij}$ : The normal base at position $i$ and depth $d$ \\
        $T_{ij}$ : The tumor base at position $i$ and depth $d$ \\        
    \columnbreak
        
    \begin{figure}[H]
      \centering
      \includegraphics[width=0.2\linewidth]{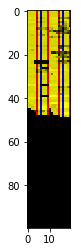}
      \caption{A visual representation of the $pair$ matrix with $w=10$ and $d=100$}
      \label{fig:boat1}
    \end{figure}
        
\end{multicols}

\subsection{The Labels}

For Training, we used the Multi-Center Mutation Calling in Multiple Cancers (MC3) high confidence variant dataset \cite{MC3paper}. The labels for this dataset are binary: 1 if the variant is validated positive, 0 if the variant is validated negative. We did not include non-validated data.

The input data is the aforementioned $pair$ matrices with a specific size of (100, 20) with three color channels. We reshape the input data from (100, 20, 3) to (100, 60), diluting the color information to analyze using LSTMs. Secondly, we balanced the dataset by undersampling the majority class. This resulted in a total of 103,868 (image, label) pairs. For these data we used a 0.8 training-test split.

\section{Methods}

\begin{figure}[H]
    \centering
    \includegraphics[width=0.9\linewidth]{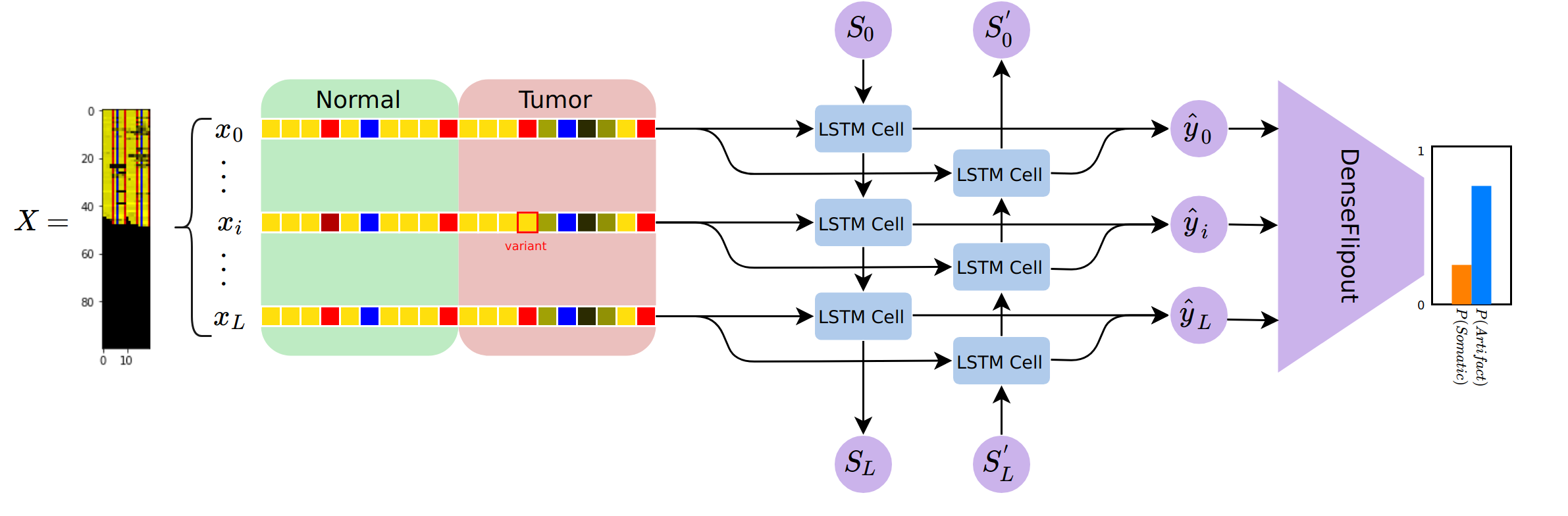}\\
    \label{fig:architecture}
    \caption{Architecture of the network}
\end{figure}

We start from the Bayesian Neural Network architecture demonstrated in Figure \ref{fig:architecture}, implemented using Tensorflow and the Tensorflow Probability library. To test its robustness and performance we compare it with a standard neural network, identical with the exception of substituting the variational layer DenseFlipout with a Dense layer.

These methods build on much of our previous work on a suitable cost function for this type of data and this classification problem\cite{DubourgFelonneau2019SafetyAR}.

We use variational layers during training with data $\mathcal{D}$, to learn an approximate probability density function $Q_{\vec{w}}(\vec{\theta})$ of the true, usually intractable, posterior $p(\vec{w}|\mathcal{D})$ for the weights of the network. This formally should be done by minimizing the KL divergence between the two. Minimizing this KL divergence is equivalent to minimizing another cost function, known as -ELBO (Evidence Lower Bound)\cite{weightuncertainty}

\begin{equation}
    -ELBO = -\mathbf{E}_{Q_{\vec{w}}(\vec{\theta})}log(P(\mathcal{D}|\vec{w}))+\mathbf{KL}(Q_{\vec{w}}(\vec{\theta})||p(\vec{w}))
    \label{eq:elbo}
\end{equation}{}

where $P(\mathcal{D}|\vec{w})$ and $p(\vec{w})$ are the likelihood of the data and the prior respectively for the weights $\vec{w}$, and $Q_{\vec{w}}(\vec{\theta})$ is an approximate pdf to the true posterior.

Practically, as the integrals in \ref{eq:elbo} are in general analytically intractable, they are computed via Monte Carlo (MC) sampling from the approximate pdf $Q_{\vec{w}}(\vec{\theta})$. This is automatically done by the variational layers in tfp. 

Secondly, as it is costly to calculate the ELBO for large datasets such as those in DNA sequencing, tfp allows to do minibatch optimization. Therefore, \ref{eq:elbo} is calculated as an average over many ELBOs calculated on batches of data $\mathcal{D}_k$ with sampled weights $\vec{w}_i$

\begin{equation}
    \hat{ELBO} = \frac{1}{N_{MC, elbo}}\sum_{i=1}^{N_{MC,elbo}}[-(log(p(\vec{w}_i))+log(p(\mathcal{D}_k)|\vec{w}_i)))+log(q(\vec{w}_i)|\vec{\theta})]
\end{equation}{}

This approximation is an unbiased stochastic estimator for the ELBO\cite{weightuncertainty}. As a result, the calculated gradient will not be miscalculated, allowing convergence to the correct solution.

Flipout estimator \cite{flipout} provides the unbiased ELBO estimation with lower variance than other methods (although larger numerical cost). This is well appreciated, since the optimization of ELBO depends largely on its variance.  

Once the pdf $Q_{\vec{w}}(\vec{\theta}) \sim p(\vec{w}|\mathcal{D})$ is fitted by minimizing -ELBO, one can infer $y^*$ with test datum $X^*$, by computing:

\begin{equation}
    p(y^*) = \int d\vec{w} p(y^*|X^*, \vec{w})p(\vec{w}|\mathcal{D})
\end{equation}{}

In practice, this is done by Monte Carlo sampling. We substitute the integral with an averaged sum over the weights sampled from the approximate posterior, then we take the average.

\begin{equation}
\hat{p}(y^*) = \frac{1}{N_{MC}}\sum_{\vec{w}_{k} \sim  Q_{\vec{w}}(\vec{\theta})} p(y^*|X^*, \vec{w}_k) \label{}
\end{equation}{}

In a classification problem, this means that for each possible outcome $y^* \in \mathcal{A}$, where $\mathcal{A}$ is the set of all outcomes, we will calculate this estimate $\hat{p}(y^*)$, ensuring that $\sum_{y^* \in \mathcal{A}} \hat{p}(y^*) = 1$.

As we will show later, the nice property of this implementation of a BNN, is that it becomes more uncertain to out of distribution samples with respect to the standard network.

As it is known, in particular for very deep one, standard neural networks suffer from uncalibration of the probability that comes from the logits. A way to overcome partially this is temperature scaling \cite{temp}.

We applied temperature scaling \cite{temp} calibration on the standard network in order to make the comparison more fair. Temperature scaling consists of minimizing the cross-entropy, after the network is trained, by fitting a new parameter $T$ with validation data, defined as follows:
\begin{equation}
\hat{y} = \frac{ e^{logits_i/T} }{ \sum{logits_j/T} }
\end{equation}{}

\begin{figure}[H]
    \centering
    \includegraphics[width=0.65\linewidth]{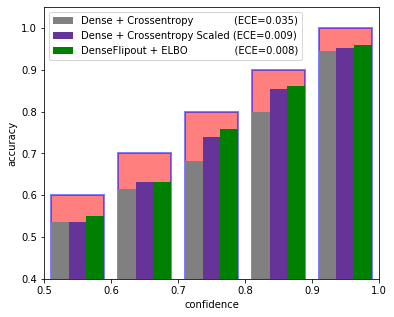}\\
    \caption{Average accuracy per confidence bin and expected calibration error (ECE)}
    \label{fig:frequency_test}
\end{figure}

In order to evaluate the calibration of a network we use Expected Calibration Error (ECE) \cite{temp}, the results of which are displayed in Figure \ref{fig:frequency_test}. This is done by sorting the predictions and partitioning them in $K$ bins. We compute, per bin, the absolute difference between the accuracy and the confidence, weighted by the bin probability.
\begin{equation}
ECE = \sum_{i=1}^K \frac{|B_i|}{n} \cdot \abs{ \frac{\sum{\hat{Y}_{i} = Y_{i}}}{|B_i|} - P_{i}}
\end{equation}{}

where $B_i$ is the $i^{th}$ bin, $n$ the number of examples, $Y_i$, $\hat{Y}_i$, $C_i$ the label, prediction and confidence vector for bin $i$.

\section{Results}

\begin{figure}[H]
    \centering
    \includegraphics[width=0.95\linewidth]{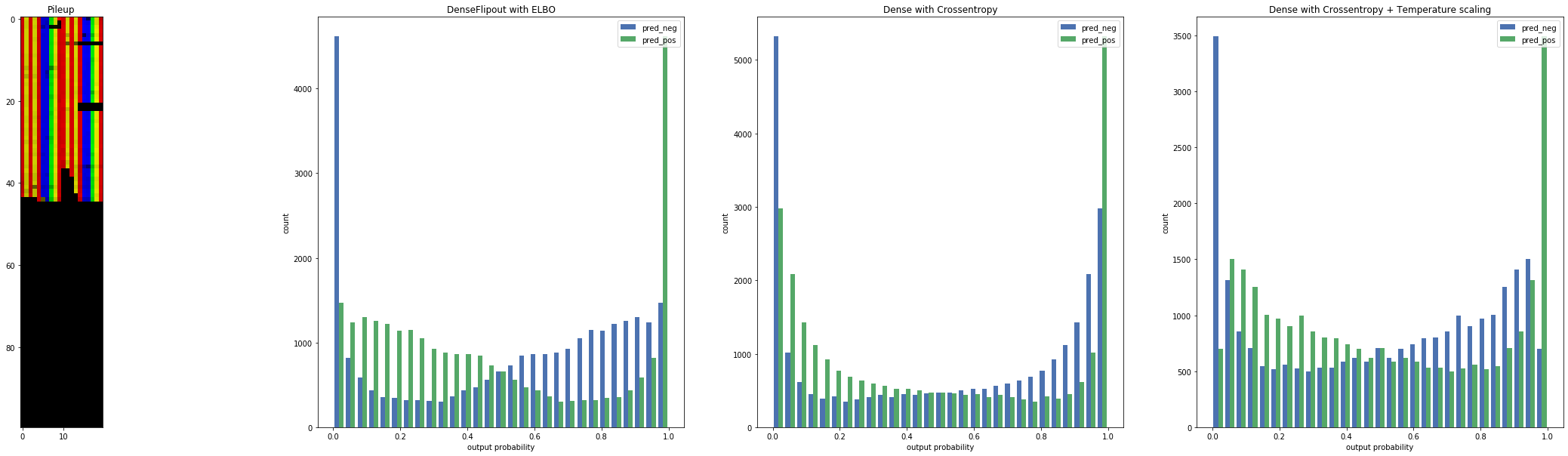}\\
    \caption{Output distributions for different models. The left-most panel displays one input $pair$ matrix. Then, from left to right, outputs are shown for the Variational Dense + ELBO, the Dense + Cross-entropy, and the Dense + Cross-entropy with temperature scaling, respectively.}
    \label{fig:bnnfig}
\end{figure}

The two models have a similar accuracy (Variational: 80\% - Dense: 81\%), although we observe a large difference in their output distributions. The variational inference outputs greater density around 0.5 than the standard network. A similar output is achieved with temperature scaling (Figure \ref{fig:bnnfig}).

\begin{figure}[H]
    \centering
    \includegraphics[width=0.95\linewidth]{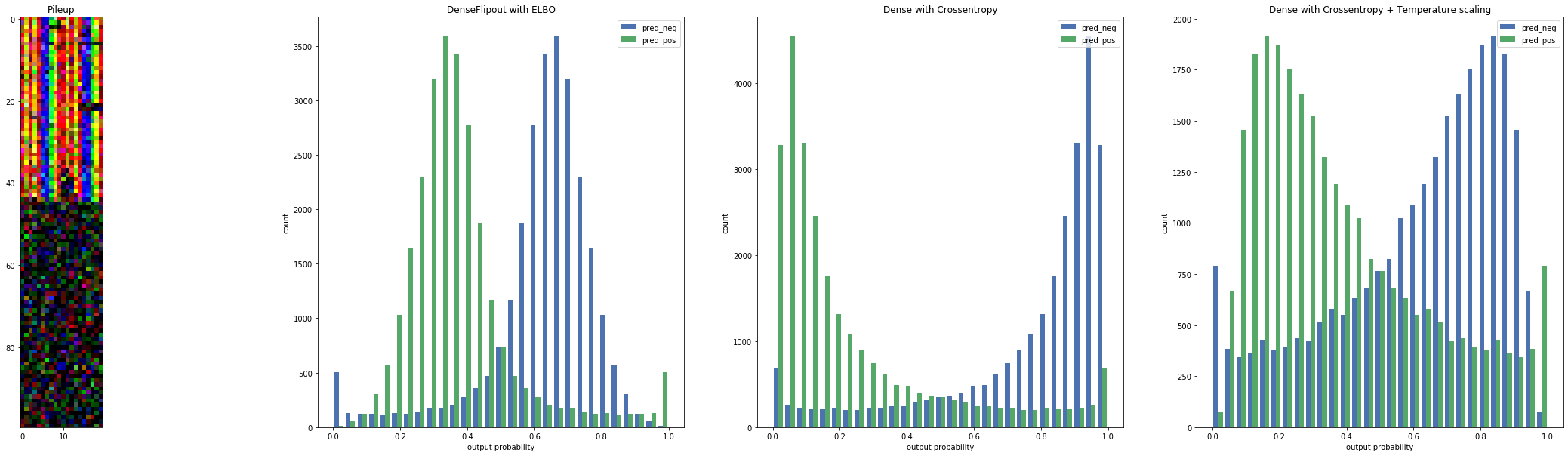}\\
    \caption{The same output distributions, for data with Gaussian noise added. The left-most panel displays an example $pair$ matrix with Gaussian noise.}
    \label{fig:bnnfig-noise}
\end{figure}

Intuitively, if the distribution of the testing data is too different from the distribution of the training data (e.g. batch variation, such as samples with a different distribution of sequencing error), we expect the confidence of the model to decrease accordingly. Such a behavior is not observed in common neural networks. To show this, we generated out-of-distribution testing data by adding Gaussian noise to the $pair$ input. We can observe (Figure \ref{fig:bnnfig-noise}) that the standard neural network is over-confident, whereas the variational approach shows a greater tendency to reduce the overall scatter between positive and negative predictions than the classic approach.

\begin{figure}[H]
    \centering
    \includegraphics[width=0.95\linewidth]{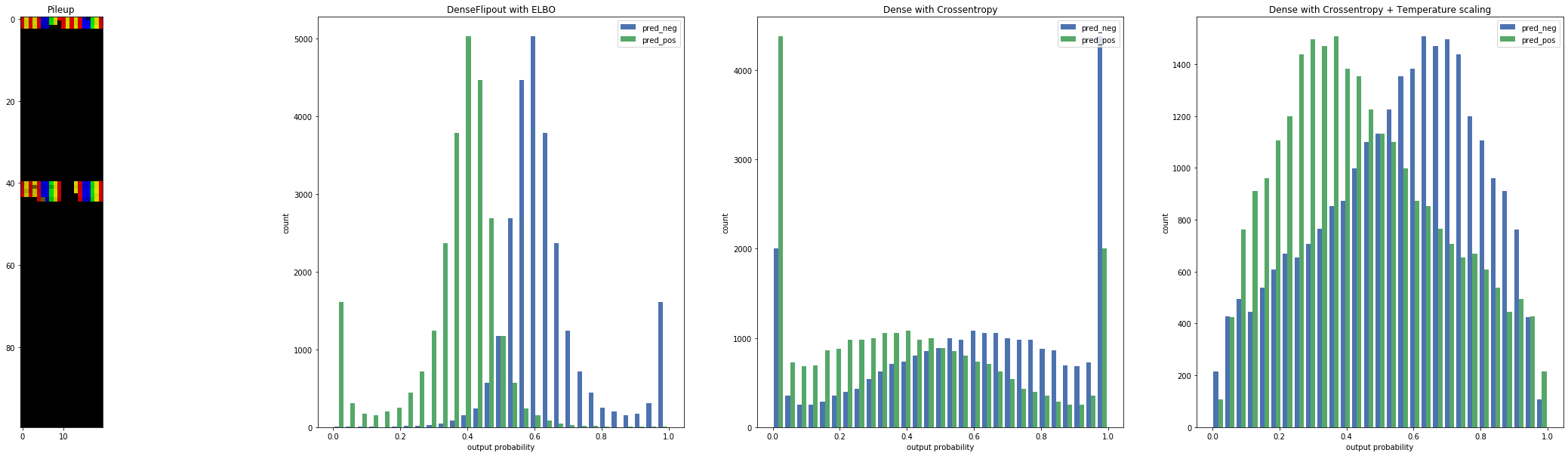}\\
    \caption{The same distributions of output probabilities, for limited data. The left-most panel displays an example $pair$ matrix with reduced sequencing depth.}
    \label{fig:bnnfig-mask}
\end{figure}

Similarly, differing sequencing depth is another commonly-occurring example of sample variability. If the network is used in a clinical setting on data with significantly lower depth than the training data, it is important for the confidence to be proportionally lowered. This is reflected in the variational model (Figure \ref{fig:bnnfig-mask}), with modes closer to 0.5 and a smaller variance than standard neural networks.

Overall, these results suggest that the variational model is best suited to disparate and highly-variable genomic sequencing data-sets, providing output probabilities that reflect the test data well.

\section{Conclusion}

We have shown in this paper that, for the application of somatic variant calling in cancer, not only can BNNs be utilized instead of standard neural networks without significant performance degradation, but also for obtaining more representative probability values for calls. Specifically, in regions where data is sparse or out-of-distribution, the network becomes uncertain and reflects this well in the output scatter. This does not occur in standard NNs, where the network outputs values of high confidence regardless of out-of-distribution scenarios. We aim to deploy these models to enable flexible but robust molecular profiling of tumors, for superior application of targeted therapies in precision oncology.

\bigskip

\bibliography{bib}

\begin{thebibliography}{11}
\providecommand{\natexlab}[1]{#1}
\providecommand{\url}[1]{\texttt{#1}}
\expandafter\ifx\csname urlstyle\endcsname\relax
  \providecommand{\doi}[1]{doi: #1}\else
  \providecommand{\doi}{doi: \begingroup \urlstyle{rm}\Url}\fi

\bibitem[Bungartz et~al.(2018)Bungartz, Lalowski, and
  Elkin]{Bungartz2018MakingTR}
Kathryn~D Bungartz, Kristen Lalowski, and Sheryl~K Elkin.
\newblock Making the right calls in precision oncology.
\newblock \emph{Nature Biotechnology}, 36:\penalty0 692--696, 2018.

\bibitem[Cibulskis et~al.(2013)Cibulskis, Lawrence, Carter, Sivachenko, Jaffe,
  Sougnez, Gabriel, Meyerson, Lander, and Getz]{Cibulskis2013SensitiveDO}
Kristian Cibulskis, Michael~S. Lawrence, Scott~L. Carter, Andrey~Y. Sivachenko,
  David~B. Jaffe, Carrie Sougnez, Stacey~Bolk Gabriel, Matthew~L Meyerson,
  Eric~S. Lander, and Gad Getz.
\newblock Sensitive detection of somatic point mutations in impure and
  heterogeneous cancer samples.
\newblock In \emph{Nature Biotechnology}, 2013.

\bibitem[Saunders et~al.(2012)Saunders, Wong, Swamy, Becq, Murray, and
  Cheetham]{Saunders2012StrelkaAS}
Christopher~T. Saunders, Wendy S.~W. Wong, Sajani Swamy, Jennifer Becq, Lisa~J.
  Murray, and R.~Keira Cheetham.
\newblock Strelka: accurate somatic small-variant calling from sequenced
  tumor-normal sample pairs.
\newblock \emph{Bioinformatics}, 28 14:\penalty0 1811--7, 2012.

\bibitem[Poplin et~al.(2018)Poplin, Chang, Alexander, Schwartz, Colthurst, Ku,
  Newburger, Dijamco, Nguyen, Afshar, Gross, Dorfman, McLean, and
  DePristo]{Poplin2018AUS}
Ryan Poplin, Pi-Chuan Chang, David Alexander, Scott Schwartz, Thomas Colthurst,
  Alexander Ku, Dan Newburger, Jojo Dijamco, Nam Nguyen, Pegah~Tootoonchi
  Afshar, Sam Gross, Lizzie Dorfman, Cory~Y McLean, and Mark~A. DePristo.
\newblock A universal snp and small-indel variant caller using deep neural
  networks.
\newblock \emph{Nature Biotechnology}, 36:\penalty0 983--987, 2018.

\bibitem[Dubourg-Felonneau et~al.(2019{\natexlab{a}})Dubourg-Felonneau,
  Rebergen, Parsons, Thompson, Cassidy, Patel, and Clifford]{ESMOnet}
G~Dubourg-Felonneau, D~Rebergen, C~Parsons, H~Thompson, J~W Cassidy, N~Patel,
  and H~W Clifford.
\newblock 1438psomaticnet: Neural network evaluation of somatic mutations in
  cancer.
\newblock \emph{Annals of Oncology}, 30, 2019{\natexlab{a}}.
\newblock \doi{10.1093/annonc/mdz257.033}.

\bibitem[Harries et~al.(2018)Harries, Zhang, Dubourg-Felonneau, Farmery, Sinai,
  Taylor, Patel, Cassidy, Shawe-Taylor, and Clifford]{Harries2018}
Luke~R. Harries, Suyi Zhang, Geoffroy Dubourg-Felonneau, James H.~R. Farmery,
  Jonathan Sinai, Belle Taylor, Nirmesh Patel, John~W. Cassidy, John
  Shawe-Taylor, and Harry~W. Clifford.
\newblock Interlacing personal and reference genomes for machine learning
  disease-variant detection.
\newblock \emph{ArXiv}, abs/1811.11674, 2018.

\bibitem[Dubourg-Felonneau et~al.(2019{\natexlab{b}})Dubourg-Felonneau,
  Darwish, Parsons, Rebergen, Cassidy, Patel, and
  Clifford]{DubourgFelonneau2019SafetyAR}
Geoffroy Dubourg-Felonneau, Omar Darwish, Christopher Parsons, Dami Rebergen,
  John~W. Cassidy, Nirmesh Patel, and Harry~W. Clifford.
\newblock Safety and robustness in decision making: Deep bayesian recurrent
  neural networks for somatic variant calling in cancer.
\newblock 2019{\natexlab{b}}.

\bibitem[Ellrott et~al.(2018)Ellrott, Bailey, Saksena, Covington, Kandoth,
  Stewart, Hess, Ma, Chiotti, McLellan, et~al.]{MC3paper}
Kyle Ellrott, Matthew~H Bailey, Gordon Saksena, Kyle~R Covington, Cyriac
  Kandoth, Chip Stewart, Julian Hess, Singer Ma, Kami~E Chiotti, Michael
  McLellan, et~al.
\newblock Scalable open science approach for mutation calling of tumor exomes
  using multiple genomic pipelines.
\newblock \emph{Cell systems}, 6\penalty0 (3):\penalty0 271--281, 2018.

\bibitem[C. et~al.()C., J., K., and D.]{weightuncertainty}
Blunderll C., Cornebise J., Kavukcuogiu K., and Wierstra D.
\newblock Weight uncertainty in neural networks.

\bibitem[Wen et~al.()Wen, Vicol, Ba, Tran, , and Grosse]{flipout}
Yeming Wen, Paul Vicol, Jimmy Ba, Dustin Tran, , and Roger Grosse.
\newblock Flipout: Efficient pseudo-independent weight perturbations on
  mini-batches.

\bibitem[Guo et~al.(2017)Guo, Pleiss, Sun, and Weinberger]{temp}
Chuan Guo, Geoff Pleiss, Yu~Sun, and Kilian~Q. Weinberger.
\newblock On calibration of modern neural networks.
\newblock In \emph{Proceedings of the 34th International Conference on Machine
  Learning - Volume 70}, ICML'17, pages 1321--1330. JMLR.org, 2017.
\newblock URL \url{http://dl.acm.org/citation.cfm?id=3305381.3305518}.

\end{thebibliography}

\end{document}